# Predictive Maintenance – Bridging Artificial Intelligence and IoT

Gerasimos G. Samatas, Seraphim S. Moumgiakmas, George A. Papakostas*
*HUman- Machines INteraction Laboratory (HUMAIN-Lab), Department of Computer Science,*
*International Hellenic University*
Kavala, 65404 Greece
{gesamat, semoumg, gpapak}@cs.ihu.gr

*Abstract*—**This paper highlights the trends in the field of predictive maintenance with the use of machine learning. With the continuous development of the Fourth Industrial Revolution, through IoT, the technologies that use artificial intelligence are evolving. As a result, industries have been using these technologies to optimize their production. Through descriptive research conducted for this paper, conclusions were drawn about the trends in Predictive Maintenance applications with the use of machine learning bridging Artificial Intelligence and IoT. These trends are related to the types of industries in which Predictive Maintenance was applied, the models of artificial intelligence were implemented, mainly of machine learning and the types of sensors that are applied through the IoT to the applications. Six sectors were presented and the production sector was dominant as it accounted for 54.54% of total publications. In terms of artificial intelligence models, the most prevalent among ten were the Artificial Neural Networks, Support Vector Machine and Random Forest with 27.84%, 17.72% and 13.92% respectively. Finally, twelve categories of sensors emerged, of which the most widely used were the sensors of temperature and vibration with percentages of 60.71% and 46.42% correspondingly.**

## I. INTRODUCTION

Nowadays, humanity experiences the era of the Fourth Industrial Revolution. This is a name given to describe the production transition from traditional ways to more autonomous and automatic. A common feature among them is the different methods used for production thus implementing radical technology improvements. Following this description, the latter-day digitalization of production deems to have the same characteristics compared to the other three Industrial Revolutions [1]. The aforementioned revolutions, also called Industry 1.0, 2.0, 3.0, and 4.0. Their evolution overtime is as follows:

- *Industry 1.0*: Creation of machines that used the power of water and steam.

- *Industry 2.0*: Utilization of electricity as mainspring.

- *Industry 3.0*: The transition from analogue devices to electronic and production automation.

- *Industry 4.0*: The improvement of technology to industrial production, the automation of various systems, industry's digitalization and the emerging trend of the Internet of Things (IoT) implantation [2].

The IoT is an efficient way to use various devices via their interaction [3], which is a connection establishment using various protocols. Thereby created a common communication basis for the devices that are part of the IoT network. Protocols have two major categories about the way they exchange data, publish/subscribe (PS) and request/ response (RR) [4]. The main difference between these two categories is that the PS uses single-directional communication, while the RR allows bidirectional communication. However, the suitability of each category is judged depending on the IoT application.

The most widely used communication IoT protocols applied in the field of PdM using machine learning are MQTT, SMQTT, AMQP, CoAP, XMPP, DDS, RESTful and WebSocket. Using the above advanced connection technologies, many data containing useful information are created. This fact along with the industry's need to reduce operational costs has driven the industry owners to use new technologies for the maintenance of their assets [5].

There are three categories by which the maintenance in general consists, considering its application time: *unscheduled*, *scheduled* and *predictive* maintenance. In the literature they are also known as Run-to-Failure(R2F), Preventive Maintenance (PvM) and Predicted Maintenance (PdM), respectively.

- **R2F**: The maintenance activities take place after a component's failure. This way is the most efficient one for cost reduction, because resources don't get wasted. On the other hand, this type of maintenance has a lot of risks, because a failure must occur prior to the maintenance. Moreover, sometimes these failures are critical, like an airplane's engine failure, for example.

- **PvM**: These maintenance activities are preventive, using a pre-scheduled plan. The part for maintenance is replaced on a specific date. This fact ensures a low possibility of sudden failure for the part involved. Compared to R2F, PvM provides more safety due to the fact that a part failure is not mandatory prior to maintenance. But this is not cost-efficient enough, because some parts will be functional after the removal so the replacement wasn't necessary.

- **PdM**: This is the maintenance of the components that really need maintenance. To achieve this prediction, the part must be continuously monitored by various sensors and simultaneously gathering and saving many data. The data are forwarding to a state-of-the-art maintenance prediction model and the model predicts on an Ad hoc basis the part replacement. PdM surpasses the other categories. It is safer from the R2F because there is no failure to any part. Furthermore, it is more cost-efficient than PvM, due to the fact the replacement concurs only to the parts that will have a failure soon, based on model prediction [6].

Modern techniques for feature extraction include *vibration-acoustic analysis*, *infrared monitoring* and *model-based condition analysis*. Various sensors are installed in the device where PdM applies without affecting the operation of the main component, which is very important because nowadays the interruption of a production line is quite costly. Indicatively in the United States, the annual cost of additional maintenance and reduced production is estimated at $750 billion [7].

The current trend in data processing and analysis is the use of *Machine Learning* (ML) methods. Machine learning has evolved into a powerful tool that implements high quality intelligent algorithms for prediction. A feature of ML is the ability to process big data and therefore effectively find hidden correlations between them. These data can be complex and produced in constantly changing environments. Such environments are the heavy industry, the automotive industry, aviation/aeronautics, business assets and even various household appliances. In each of them, of course, when PdM is applied, the economic benefits are great [8].

The current work presents a systematic literature review. Its contribution is to highlight through research the areas where PdM is applied most often, the most common measurement sensors, as well as the most prevalent machine learning models applied to PdM, thus bridging Artificial Intelligence with IoT for PdM. The challenges of predictive maintenance are also presented. More specifically, the problem of obtaining quality data for the implementation of a problem is discussed. Also, the advantages and disadvantages of using artificial intelligence in real world problems are pointed out.

The paper is organized as follows: Section II describes the way the various ML models are implemented to solve a real-word PdM problem. Section III, presents the bibliography selection protocol applied to analyse the literature and illustrates the statistics of the analysis and the taxonomy of the studied papers. Finally, Section IV summarizes the challenges of PdM with ML methods followed by Section V that concludes this study.

## II. PREDICTIVE MAINTENANCE USING MACHINE LEARNING

When ML is applied to PdM, follows a process consisting of specific stages. The aim is to achieve successful prediction for maintenance through the training of models, with more efficient data collection and use. About the data gathering, this process consists of the stage of selection and pre-processing of historical data. Historical data is the data that have been recorded throughout past working cycles. The data are collected through sensors that are part of an IoT network tailored to the system and then they are constantly updated through new measurements Fig. 1. Depending on how they were collected and stored, valuable data are exported for application to an ML model. After that they are pre-processed in order to transform and obtain the format in which it will be efficient for the model. The process continues with the selection, training and validation of the model Fig. 2. This has as a result the most appropriate ML model to be selected in order to be applied in PdM.

A fully training cycle of the ML model is completed when the data collected in the initial stages of the process, the validation of the training concurs and finally when the evaluation of the model is taking place in order to check its efficiency and effectiveness. The last stage of the process is the maintenance of the model. With this step, after selecting the appropriate algorithm, a procedure is followed in order to maintain the performance of the model in the long run and to renew its database. Fig. 3. shows a typical way of operating an ML model in a PdM application, following the above steps.

**IoT**

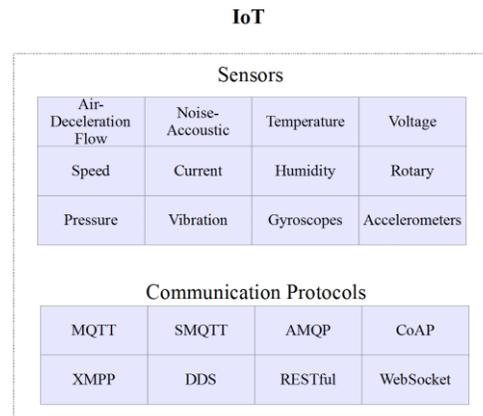

Fig. 1. IoT sensors and Protocols

**Artificial Intelligence**

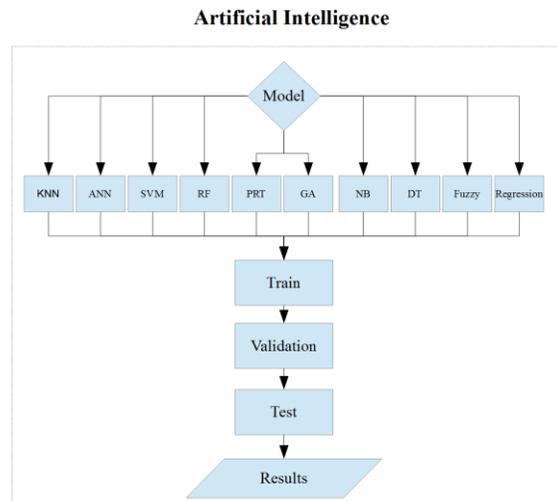

Fig. 2. AI models

**Predictive Maintenance using Artificial Intelligence**

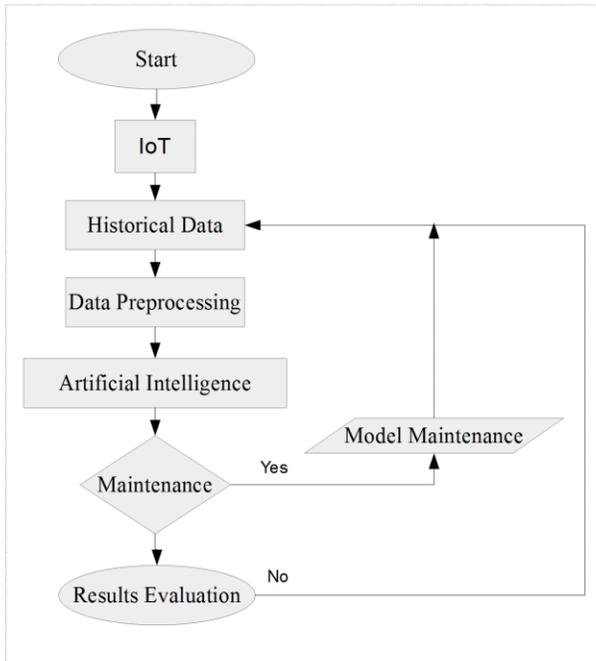

Fig. 3. PdM using AI Flowchart

The key factor for the effectiveness of ML models in PdM applications is the IoT network. Through this network, the sensors provide measurements which are processed and distributed in the database in order to be used for maintenance prediction through the model. The exchange of data within the network is done with the help of protocols that are implemented in order to achieve communication between the various sensors and systems. IoT is the basis for the efficient use of ML in PdM as without it would be impossible to receive and store multiple data from the sensors. Furthermore, the stage of selecting the historical data and the maintenance of the model, are the basics principles that differentiate the common ML models from the models intended for PdM. The way of pre-processing the data, the model selection, training, and validation are following the form of common machine learning models' corresponding stages.

## III. LITERATURE ANALYSIS

This section describes the literature analysis performed. More specifically, the research parameters and criteria used in order to select the appropriate papers and taxonomy of the papers are provided.

### A. Research Execution

The Systematic Literature Review (SLR) used in this paper is a scientific method, widely used to review various topics of interest. It is secondary research, as it takes into account other scientific research related to the same subject. After collecting the data, they evaluated and finally ends up with analysis [9]. For the bibliography selection, a protocol was applied, which was a combination of criteria with qualitative and quantitative characteristics. More specifically, the following criteria were set from the beginning of the research:

*1) Research Questions*
  *a) E1: How does PdM relate to ML?*
  *b) E2: In which sectors PdM is applied in the years 2011-2020?*
  *c) E3: By what methods was PdM applied in the years 2011-2020?*
  *d) E4: Which sensors were most applied in PdM in the years 2011-2020?*
  *e) E5: What is the trend in PdM ML methods?*

*2) Research Database*
  *To extract a lot of data, the research was done in many databases from the website Google Scholar which contains 90% of the scientific publications written in English. The tool, which used to group and export the data in editable format, was Harzing's Publish or Perish (HPP).*

*3) Rejection Criteria*
  *The following criteria were set during the execution of the research*
  *a) K1: Publications must be from the last decade from the year 2011 until 2020*
  *b) K2: Citations per year must be over 4*
  *c) K3: Total citations must be over than 10*

*4) Quality Criteria*
  *K4: The scientific publication must refer to PdM*

*5) Acceptance Criteria*
  *a) K5: Scientific publications must indicate the PdM sectors applied.*
  *b) K6: Scientific publications must indicate the machine learning method used.*

Based on the above criteria and questions for best results prior to creating the report, a query was provided to the HPP database. A time period from 2011 to 2020 was selected. The word Predictive Maintenance was placed as the title for the search and the word Machine Learning was used as the keyword. Difficulties were encountered in finding the right words to be as a title or keywords. The problems arose mainly from the fact that PdM pre-existed without its mandatory implementation with ML. So, without the term Machine Learning as a keyword, the posts were for PdM, but it was not always combined with ML. Also, the term Predictive Maintenance became mandatory in its existence in the title of the publication. This was preferred because in many publications the word PdM was a mere reference to a simple ML domain, but without much detail. The goal from the beginning was to find PdM publications implemented with ML, in order to draw appropriate conclusions.

The execution of the above conditions took place in HPP on December 10, 2020, with a limit of 1000 results. At the end of the search, the data were integrated into a worksheet and filters were applied in order to group them. The total publications were 847. Initially, posts with zero citations were excluded and 483 posts were generated. Thus, arose the need to create criteria K2 and K3. The following method was used to create them. The average of the remaining reports was calculated and the number 10.7 was obtained. Thus, K2 had a value higher than 10 and when it was implemented, 115 publications emerged. The value of K3 was calculated in the same way with the average of the reports being 4.2. When K3 was implemented, the posts were 89. K4, K5 and K6 were applied to them, which driven to 12 more publications rejection. Following the above procedure, finally, there were 44 publications with a PdM model implementation and 33 with methods analysed.

### B. Taxonomy

This paragraph taxonomizes the papers into three groups according to the sectors in which PdM had been applied, the most frequently used ML methods and the various sensors described in the papers. A thorough study of the publications was required to extract the data and as a result, the data came only from the 44 publications with a PdM model implementation.

The first taxonomy relates to the sectors where PdM using Machine Learning applied. Through this taxonomy, six categories emerged. These are Aviation/Aeronautics, Business, Electronics, Production, Energy and Transportation. The field of Aviation and Aeronautics [10]-[12] includes implementations related to the maintenance of flying vehicles, aircraft and helicopters. Business [13]-[15] includes topics related to business studies such as damage to a company building, damage to a copier, or even the freezer of a grocery store. PdM implementations in various electronic components are reported as Electronics [16]-[20]. Production [21]-[44], referred to machines that a factory needs to produce any kind of a product. Moreover, the Energy [45]-[49] refers to systems related to energy production and finally the term Transportation [50]-[53] is used for transportation PdM implantation.

The second way to taxonomize the papers was done according to the ML model that was applied. Here the issue that arose, was that there were several different models with many subcategories. Thus, in order to create a reliable result, the methods were also grouped. The first category is the Artificial Neural Network (ANN) which contains all the categories of Neural Networks. The second is the Support Vector Machine (SVM), then the Random Forest, followed by K-Nearest Neighbour (KNN), Decision Tree, Regression Analysis (which included logistic and linear regression) and finally the Naïve Bayes networks (which includes the classic Naïve Bayes and the Hidden Markov Model, which have a similar philosophy), the Fuzzy Systems, the Genetic Algorithms and finally the Pattern Recognition Techniques (PRT).

Finally, the third group was about the various sensors used. More than one sensor was used in each application. However, twelve categories of sensors appeared based on their purpose and function. Sensor categories are Temperature, Vibration, Noise /Earphone, Pressure, Accelerometer, Rotary, Air Flow, Deceleration, Humidity, Gyroscope, Current, Speed and Voltage. Category names were given from the function of each sensor applied to the PdM.

### C. Statistics

This section contains statistical information on the data extracted from the previous section. Starting from the sectors which PdM using Machine Learning applies, the 6 categories were grouped by descending order of publications and the order is: Production, Electronics, Energy, Transport, Business and Aviation/Aeronautics. This order led to the fact that PdM in the Production sector has greater exposure to ML methods, compared to other sectors with a percentage of 48%, thus demonstrating the great ML impact in this sector. In fact, this high percentage was important because the other sectors had lower values such as 11%, 11%, 9%, 7%, 7% and 7% accordingly. As a result, it became clear that the vast majority of PdM applications with ML in the years 2011-2020 were implemented in Production. This fact strengthens the hypothesis that PdM, offers substantial profits to its operators and especially for sectors such as production where a sudden failure in a machine, can cause irreparable damage with significant economic consequences.

The next statistics are for the ML methods applied in PdM. As mentioned before, grouping was performed between the different methods, and as a result the following ten categories emerged grouped by descending order of publications: ANN, SVM, Random Forest, KNN, Decision Tree, Regression Analysis, Naïve Bayes, Fuzzy Systems, Genetic Algorithms and PRT. First in line is the ANN and next SVM and Random Forest. The percentages were 29%, 18% and 15% respectively. These ML methods were followed by KNN, Decision Tree and Regression Analysis with percentages of 12%, 9% and 7%. Then there were Naïve Bayes, Fuzzy Systems and Genetic Algorithms at 5%, 3% and 1%. Finally, the PRT had the last position with a 1% percentage. It was observed that except for the ANN, the other methods had the same differences with each other, ranging from 2 to 3 percent. The ANNs from the second SVM had a difference of 11%. This difference may not be large enough but it remains significant. Also, the first two methods covered 47% of the cases while the rest of the percentage was divided into the other eight models. With the third method Random Forest, taking account then the percentage of the three dominated methods increased to 62%.

Data are fundamental parts of an ML PdM application. The data are generated when measurements are taken from the sensors, the machines, the systems, or even a part of them. As already mentioned, based on the aforementioned scientific studies, twelve sensor categories emerged. The most commonly used sensor categories are Temperature, Vibration and Noise/Acoustic with 60.71%, 46.42% and 32.14% respectively. The percentage of the Pressure was 28.57%, of the Accelerometer 25% and of the Rotary 14.28%. The Air / Deceleration Flow and Humidity Sensors had the same percentage at 10.71%. Finally, Gyroscopes, Current, Speed and Voltage also had the same percentage which was 7.14%.

The following Fig 4, Fig5 and Fig 6 visualize the previous data to reveal the trend in each sector in PdM applications.

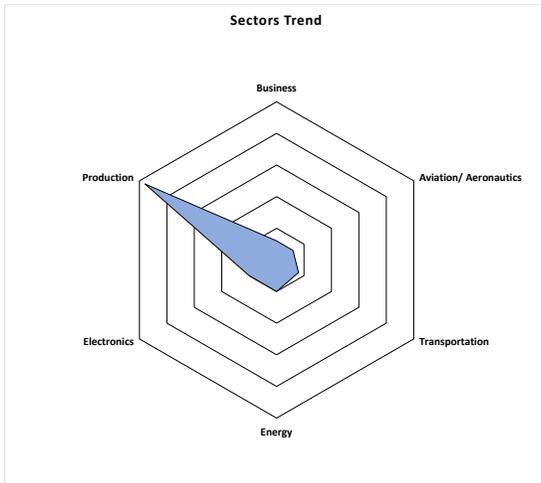

Fig. 4. Sector Trend

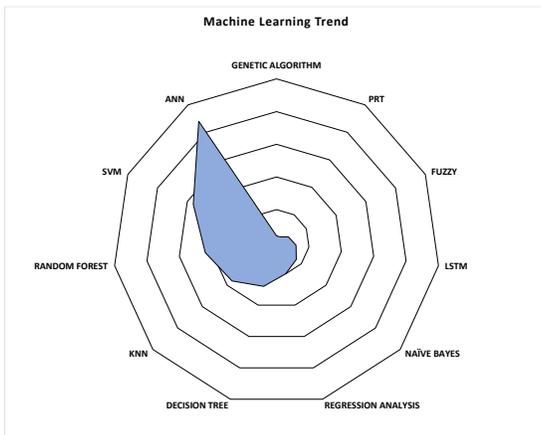

Fig. 5. Machine Learning Model Trend

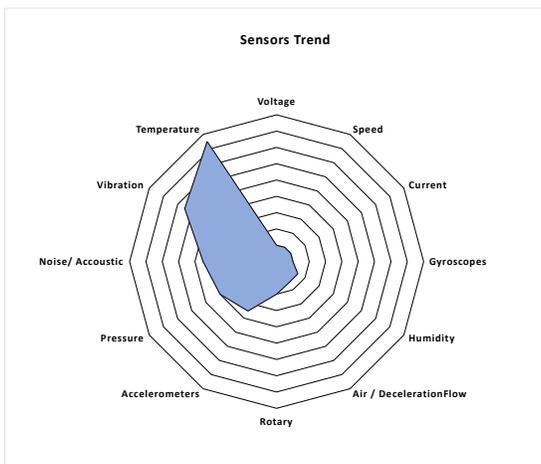

Fig. 6. PdM Sensors Trend

## IV. DISCUSSION

The datasets of all the publications studied, played an important role in the solution design of the problem. By definition, artificial intelligence learning processes are based on data, which must meet both quantitative and qualitative criteria. More specifically, a lot of data must be ensured for the training to be more effective. In the case of PdM, the data relates to the operation of the equipment. This function can include normal operation and operation with errors or only data resulting from errors. In the first case, assumed that several operating conditions have been recorded, there is a lot of data. But here comes the second criterion, the quality. The normal state of a machine is its smooth operation. When there is an error and it is recorded there will be a lot of data from normal operation and little from the fault, which creates several problems in the training of models.

This problem is referred to in the literature as *imbalanced data* and is a common problem when the datasets are from real-world problems. Usually, the class that is not numerically superior is the one that is examined containing the fault conditions. During the execution of the prediction model, 4 states are created. Two of them are when the prediction is correct for the correct and incorrect operation. The other two are when this prediction has been failed. Briefly these terms are referred to as True Positive, True Negative, False Positive and False Negative, where the positive is the normal function and negative is not. There are several approaches to solving this problem. The most common is to recalculate the sample. There are three methods, Oversampling, Subsampling and the combination between the two. Subsampling is usually done in the dominant class to reduce the size while in the less powerful class applied techniques such as data synthesis or duplicate data [54].

Furthermore, the sensors are the mean by which the data are created. As mentioned in the previous sections, they are integrated at various positions and collect several different measurements. Their accuracy is extremely important as the training of the model and the correct prediction are based on the data provided by them. Advances in their technology have created very small sensors that can be easily integrated into almost any construction. Also, the measurements, nowadays, have exceptional accuracy, which produces reliable results when creating the model. However, further improvements in technology will result in much smaller and more accurate sensors, resulting in additional sensors as well as more efficient models.

In recent years, Artificial Intelligence has developed to a great extent. Models created in conjunction with increasing data access create high-performance models. A major hurdle to overcome is that the ML models used are characterized as black-boxes. As a result, there are insufficient explanations when a system should stop working for maintenance activities. Future efforts are directed towards solving the problem and when some progress had been made a further increasing tendency to use ML methods in PdM will be shown. Moreover, nowadays the trend is increasing.

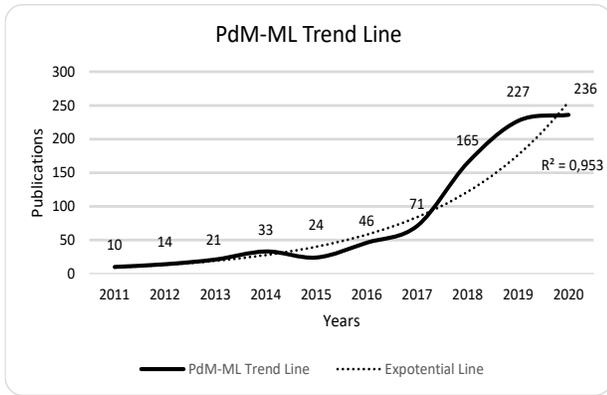

Fig. 7. PdM using ML trend Line 2011-2020

Exporting data from the early 847 papers, applying only K1 criterion, which covers the years 2011-2020, it was observed that the data show an exponential increase during the last decade. The $R^2$ index was used and showed that the trend line was almost identical to the exponential growth line. A small exception was in 2020, where there was a slight decrease in growth, but this is justified due to the specificity of 2020 which was a pandemic year and had affected, among other things, research in all areas. Fig. 7 shows this trend line.

## V. CONCLUSION

The present paper was a result of a systematic review conducted on 44 publications with PdM implementations that used ML methods over a decade. Their selection was made through criteria that are clearly mentioned in the previous sections. There was also an issue with the publications of the year 2020, which due to the criteria applied on the number of publications, in conjunction with the COVID-19 pandemic was lower than expected. The research showed the Production be the predominant sector to which PdM applies. The methods that emerged, showed that the most prevalent are ANN, SVM and Random Forest, but the choice of the appropriate one depends on the data of each problem.

Furthermore, more than one sensor was used in each application. The reason for that fact was that most PdM applications require a lot of different data and this is a way to achieve this. Moreover, twelve categories of sensors emerged the temperature, vibration, noise and acoustic sensors, which were the most used in PdM applications using ML. Based on this, the conclusion is that temperature, vibration and noise are key factors to be implemented in a PdM ML model.

Also, in the future the methods applied in PdM could be studied by sector in order to study the performance of the models in specific sectors and to draw additional conclusions. Sensors can also be associated with sectors, in order to emphasize specific sensors. Finally, there is the assessment according to the graphs presented and the fact that efforts are being made to overcome the problem of blackbox models, that in the coming years the trend line will continue to be exponential and more companies will invest in PdM and especially in the Production sector.

## ACKNOWLEDGEMENT

This work was supported by the MPhil program "Advanced Technologies in Informatics and Computers", hosted by the Department of Computer Science, International Hellenic University, Greece.